\documentclass{article}
\usepackage{ijcai23}

\usepackage{times}
\usepackage{soul}
\usepackage{url}
\usepackage[hidelinks]{hyperref}
\usepackage[utf8]{inputenc}
\usepackage[small]{caption}
\usepackage{graphicx}
\usepackage{amsmath}
\usepackage{amsfonts}
\usepackage{amsthm}
\usepackage{booktabs}
\usepackage{algorithm}
\usepackage{algorithmic}
\usepackage[switch]{lineno}
\usepackage{multirow}

\title{OrchMoE: Efficient Multi-Adapter Learning with Task-Skill Synergy}

\author{
Haowen Wang$^1$
\and
Tao Sun$^1$\and
Kaixiang Ji$^1$\and
Jian Wang$^1$\and
Cong Fan$^1$\and
Jinjie Gu$^1$
\affiliations
$^1$AntGroup\\
\emails
\{wanghaowen.whw, suntao.sun, kaixiang.jkx, bobblair.wj, fancong.fan, jinjie.gujj\}@antgroup.com
}
\begin{document}

\maketitle  

\begin{abstract}
We advance the field of Parameter-Efficient Fine-Tuning (PEFT) with our novel multi-adapter method, $\texttt{OrchMoE}$, which capitalizes on modular skill architecture for enhanced forward transfer in neural networks. Unlike prior models that depend on explicit task identification inputs, $\texttt{OrchMoE}$ automatically discerns task categories, streamlining the learning process. This is achieved through an integrated mechanism comprising an Automatic Task Classification module and a Task-Skill Allocation module, which collectively deduce task-specific classifications and tailor skill allocation matrices. Our extensive evaluations on the 'Super Natural Instructions' dataset, featuring 1,600 diverse instructional tasks, indicate that $\texttt{OrchMoE}$ substantially outperforms comparable multi-adapter baselines in terms of both performance and sample utilization efficiency, all while operating within the same parameter constraints. These findings suggest that $\texttt{OrchMoE}$ offers a significant leap forward in multi-task learning efficiency.
\end{abstract}

\section{Introduction}

The inexorable increase in the number of parameters within large language models (LLMs) has precipitated a commensurate rise in the computational resources requisite for training these models, as well as their multimodal derivatives. In this computational milieu, Parameter-Efficient Fine-Tuning (PEFT) has been acknowledged as a parsimonious fine-tuning approach. This method involves the static retention of the base model's parameters while selectively updating the parameters within the adapter layers. LoRA~\cite{DBLP:journals/corr/abs-2106-09685}, AdaLoRA~\cite{zhang2023adalora}, and IA3~\cite{liu2022few} have been substantiated as robust foundational adapters integral to the PEFT framework.

The rigorous validation and refinement of adapter-centric PEFT approaches suggest a promising frontier for addressing the challenges of multi-task model fine-tuning against the backdrop of increasingly capacious models. This proposition is founded on the collaborative training of a suite of adapters. During multi-task training, each foundational adapter is conceptualized as a discrete 'skill set,' with the amalgamation of varied foundational adapters orchestrated through a learned allocation matrix that assigns different tasks to their respective skills.

Notwithstanding the theoretical and empirical validation of this training construct by preceding research, including the Poly~\cite{ponti2022combining} and MHR~\cite{caccia2023multihead} studies, the dependency on explicit recognition of task identifiers within multi-task datasets constrains their broader implementation. Indeed, many real-world settings are either incapable of or disinclined to undertake the explicit demarcation of tasks due to logistical complexities or prohibitive costs. To circumvent these limitations, methodologies such as MoLoRA~\cite{zadouri2023pushing} and MultiLoRA~\cite{wang2023multilora} have been proposed. These methods facilitate joint optimization of multiple adapters and are capable of either distinguishing task identifiers without direct input reliance or enabling the simultaneous learning and training of multiple tasks across adapters.

To harness the capabilities of multiple adapters in addressing diverse training tasks prevalent in real-world scenarios, this study proposes $\texttt{OrchMoE}$. This approach envisages adapters as elemental skill units and distills a series of overarching tasks. In the prediction phase, mirroring human cognitive adaptability, $\texttt{OrchMoE}$ dynamically determines the requisite combination of tasks for each data sample. It further endeavors to master the process through which tasks can exploit the allocation vectors of these skills, thereby bolstering the proficiency of multiple adapters in navigating the intricacies of real-world multitasking challenges.

This article articulates several pivotal contributions to the field of Parameter-Efficient Fine-Tuning (PEFT) as follows:

\begin{itemize}

\item We propose a novel PEFT framework named $\texttt{OrchMoE}$, which is a holistic paradigm designed to synchronize multiple adapters. $\texttt{OrchMoE}$ facilitates the employment of these adapters in the fine-tuning and learning processes for a breadth of downstream tasks, thereby enhancing model versatility and adaptability.

\item The operational mechanisms of tasks and skills within $\texttt{OrchMoE}$ are demystified through detailed visualizations and theoretical explanations. This investigation delineates the means by which the Task-Skill approach incrementally refines the utilization efficiency of adapters.

\item Rigorous experimental validation was carried out using the comprehensive SuperNI dataset, which comprises 1600 instructional tasks. These experiments tested the efficacy of $\texttt{OrchMoE}$ across both Encoder-Decoder and Decoder-Only model architectures. The findings from our study demonstrably indicate that $\texttt{OrchMoE}$ substantially outperforms the prevalent multi-adapter and single-adapter baselines, maintaining consistency in parameter count across various architectural models.

\end{itemize}

\section{Preliminaries}
\label{sec:preliminaries}
The Parameter-efficient Fine-tuning (PEFT) method is widely recognized as an effective approach for training models with a large number of parameters. Consider a learning weight matrix $\mathbf{W} \in \mathbb{R}^{d \times k}$. By examining the prominent adapter architecture, LoRA, it becomes evident that the training parameter space can be constrained through a low-rank decomposition strategy. This can be mathematically represented as:
\begin{equation}
\resizebox{.91\linewidth}{!}{$
\displaystyle
h = \mathbf{W}\mathbf{x} = (\mathbf{W_0}+\Delta\mathbf{W})\mathbf{x}=(\mathbf{W_0}+\mathbf{B}\mathbf{A})\mathbf{x}=
\mathbf{W_0}\mathbf{x}+\mathbf{B}\mathbf{A}\mathbf{x}
$}
\end{equation}
Here, $\mathbf{W_0}$ denotes the pre-trained weight matrix, while $\mathbf{A}$ and $\mathbf{B}$ represent the trainable parameters, with $\mathbf{B} \in \mathbb{R}^{d \times r}$ and $\mathbf{A} \in \mathbb{R}^{r \times k}$. It is important to note that $r \ll \min(d,k)$, implies a significant reduction in the dimensionality of the trainable parameters. For the purpose of this discussion, we shall assume $k=d$ henceforth.

% \begin{equation}
%     \resizebox{.91\linewidth}{!}{$
%             \displaystyle
%             x = \prod_{i=1}^n \sum_{j=1}^n j_i + \prod_{i=1}^n \sum_{j=1}^n i_j + \prod_{i=1}^n \sum_{j=1}^n j_i + \prod_{i=1}^n \sum_{j=1}^n i_j + \prod_{i=1}^n \sum_{j=1}^n j_i + \prod_{i=1}^n \sum_{j=1}^n i_j
%         $}
% \end{equation}
In this study, we explore strategies for the integration of multiple adapters, with a particular focus on the LoRA architecture. Specifically, we consider a collection of LoRA adapters characterized by the set ${(\mathbf{A}_1, \mathbf{B}_1, r_1), \cdots, (\mathbf{A}_k, \mathbf{B}k, r_k)}$. In common practice, these adapters are typically amalgamated via a weight matrix $\mathbf{Z}$. As a result, the parameter update for the combined system can be expressed as:
\begin{equation}
\Delta \mathbf{W} = \sum_{i=1}^{k} \mathbf{w}_i \mathbf{B}_i \mathbf{A}i,
\end{equation}
where $\mathbf{w}i$ denotes the weighting coefficients for each adapter. Theoretically, this aggregation is analogous to constructing a single adapter with a higher rank, denoted as $r_{\text{larger}} = \sum_{i=1}^{k} r_i$. 

Evidence has emerged suggesting that simply increasing the rank of LoRA to inflate the parameter count does not consistently result in significant improvements to the model~\shortcite{DBLP:journals/corr/abs-2106-09685}. The creators of the LoRA framework~\cite{DBLP:journals/corr/abs-2106-09685} postulate that the intrinsic rank of $\Delta \mathbf{W}$ is quite limited; consequently, enhancing the rank of LoRA is not guaranteed to capture additional meaningful subspaces. This situation takes a different turn when we consider the integration of nonlinear elements or the disruption of equivalence and symmetry between various LoRA instances. Such alterations represent a departure from the foundational principles and mechanisms that underpin mere rank augmentation in LoRA, heralding the potential for novel paradigms within expansive parameter spaces.

Building upon this understanding, our research advances a more integrative method, paralleling the Mixture of Experts (MoE) model, to address the fusion of multiple LoRAs. We train an ensemble of "small" LoRAs in tandem and, through their synergistic combination, we transform rudimentary linear combinations into adaptive, learnable configurations. This combinatorial learning schema hinges on accurately learning and applying the allocation weights $\mathbf{w} = \{w_1, w_2, \ldots, w_k\}$ associated with each LoRA module. Invoking the MoE structure, we term these allocation weights as the "Router," a crucial mechanism for allocating inputs to the most suitable LoRA, according to the learned specifications. The Router imbues the overall model with flexibility and a refined capability for specialization, marking a significant evolution from the traditional paradigms of parameter scaling.

Upon establishing the concept of the "Router" within our framework, the most straightforward method\cite{wang2023multilora} would be to instantiate a set of learnable parameters $\mathbf{w}$, which advocates for an adaptive learning of $\mathbf{w}$. In this configuration, all inputs share a common router, denoted by $\mathbf{w}$, which dynamically adjusts to the inputs rather than remaining fixed or token-specific.

In scenarios where the dataset exhibits definitive task boundaries, the strategy of task-level routing has been implemented in the context of multi-task learning\cite{ponti2022combining,wang2023customizable,caccia2023multihead}. This technique is predicated on a routing mechanism in which $\mathbf{w}^{\tau} = f(\mathbf{t}(\mathbf{x}_i))$ indicates that inputs $\mathbf{x}$, affiliated with the same task $\tau_i$, share a common set of routing parameters $\mathbf{w}^{\tau}$. Here, the function $\mathbf{t}: \mathcal{X} \to \mathcal{T}$ categorizes input $\mathbf{x}_i$ into a corresponding task $\tau \in \mathcal{T} = \{1, 2, \ldots \}$. Furthermore, the function $f: \mathcal{T} \to \mathbb{R}^k$ encapsulates the learning process that ascertains the routing parameters for each adapter, contingent upon the task associated with the inputs. Such a method ensures that inputs correlated with the same task are managed via a unified routing protocol, thus fostering task-specific learning that adeptly addresses the distinctive characteristics of each task.

To facilitate the router's adaptation to more general data scenarios, and particularly in cases lacking clear task ID delineations, a more flexible Top-K~\cite{zhou2022mixtureofexperts} routing mechanism is utilized. This strategy restricts computation to the 'k' highest-ranked experts as identified by the router, thereby introducing sparsity and reducing computational complexity. Although the computation is dependent on the selected top 'k' experts, the memory overhead is influenced by the overall number of experts. Each input token $x_{i,j}$ is endowed with a unique set of routing weights $w^{i,j} = h(e(x_{i,j}))$. The embedding function $e: \mathcal{X} \to \mathbb{R}^d$ encodes the input token $x_{i,j}$, while the mapping function $h: \mathbb{R}^d \to \mathbb{R}^k$ utilizes a straightforward network structure, such as a single-layer Multilayer Perceptron (MLP), to convert the token embeddings into a vector of weight allocations.

\section{Methodology}
\label{sec:methodology}
\begin{figure}[H]
    \centering
    \includegraphics[width=0.5\textwidth]{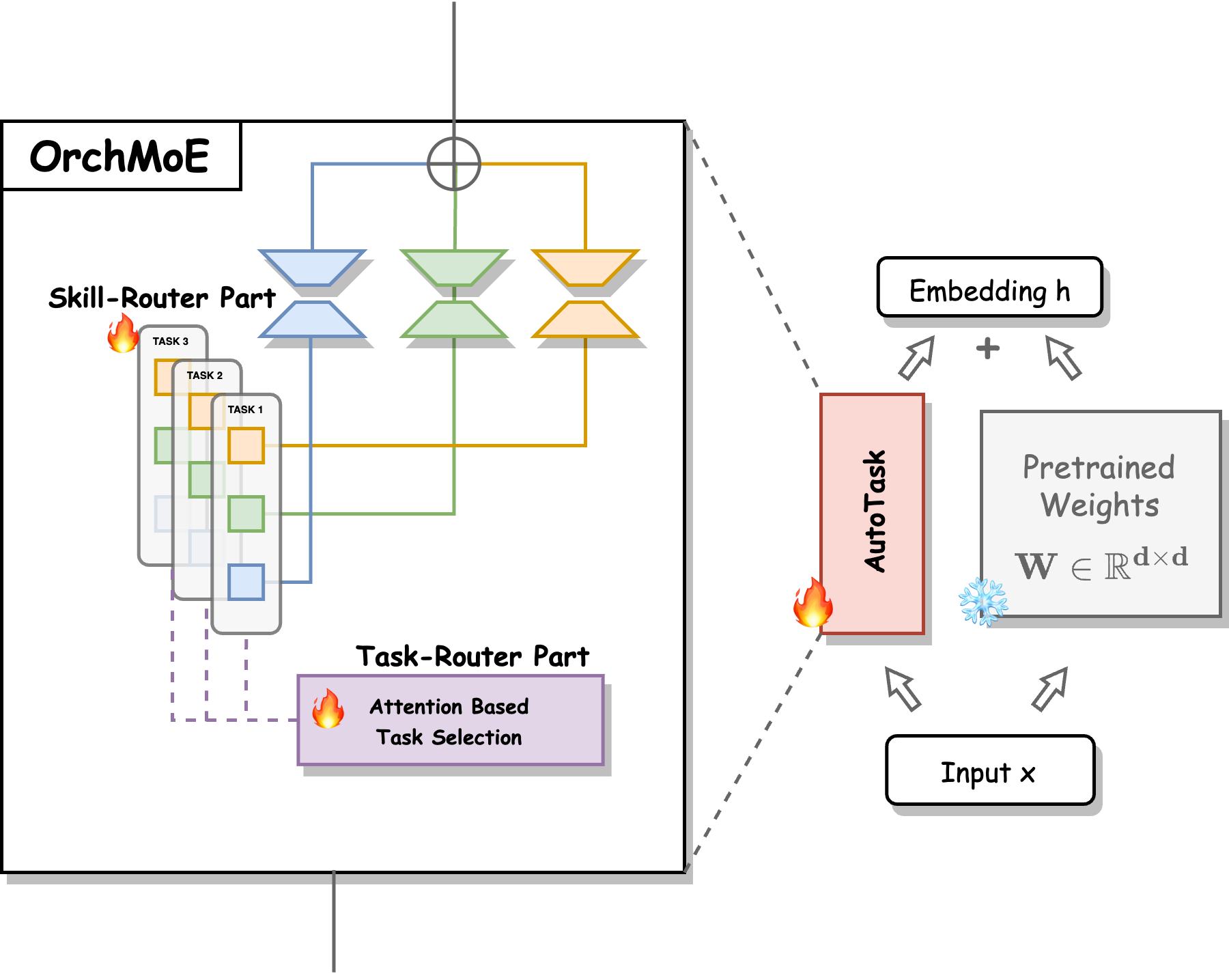}
    \caption{$\texttt{OrchMoE}$ Infrastructure}
    \label{fig:structutre_autotask}
\end{figure}

The proposed $\texttt{OrchMoE}$ strategy, as depicted in the figure~\ref{fig:structutre_autotask}, expands upon the notion of task skills within the context of multi-task PEFT (Parameter-Efficient Fine-Tuning). We envisage a scenario with $T$ abstract tasks and posit that each task can access a repertoire of skill modules denoted by $\Phi = \{\phi_1, \phi_2, \ldots, \phi_{|\Phi|} \}$. With this foundation, we construct task and skill routers predicated on the input to facilitate learning.

The task router is responsible for discerning the type of task that the current input sample is appropriately associated with. Concurrently, the skill router ascertains the optimal manner in which each task should leverage the available skills, guided by the input. This dual-router system thus dynamically aligns each input with the most suitable task classification and the most effective skill set to employ.

The overall architecture is encapsulated by the following expression:
\begin{equation}
\sum_{i=1}^{T}\sum_{j=1}^{S}w^{i}\phi_{j}^{i}x
\end{equation}

where $W$ denotes the task routing function, which determines the weighting coefficients for each task based on the given input. $\Phi$ signifies the skill routing function, essentially the learned Task-Skill Allocation Matrix that is contingent upon the characteristics of both input vectors and the corresponding tasks.

\subsection{Task Router}
The task router module encapsulates the algorithm that governs task allocation. Specifically, the function $\mathbf{w}(X)$ is employed for each input sample to assign the activation weight corresponding to each abstract task. This mechanism effectively quantifies the contribution of each task within a given sample. To augment the interaction among nonlinear features and to bolster the representational capacity of the model, we have incorporated elements from the self-attention mechanism into the architecture. The integration of self-attention components within the task router not only facilitates more intricate correlations between features but also elevates the overall efficiency of the learning process.

The task routing mechanism employs an attention-based selection method to process the inputs as delineated by the following equations:
\begin{equation}
\mathbf{w}(X) = W_{\text{Task}} \mathbf{\hat{X}} + \text{Bias}
\end{equation}
\begin{equation}
\mathbf{\hat{X}} = \text{SoftMax}\left(\frac{\mathbf{X}\mathbf{X}^\top}{\sqrt{d_k}}\right)\mathbf{X} + \mathbf{X}
\end{equation}

Here, $\mathbf{X}$ represents the input vector, and $d_k$ denotes the dimensionality of the hidden size within the input tensor. The matrix $W_{\text{Task}} \in \mathbb{R}^{d_k \times T}$ is the learned weight matrix, where $T$ corresponds to the number of abstract tasks.

\subsection{Skill Router}

The Skill router delineates the utilization of skills for each task, incorporating a base adapter characterized by a learnable matrix $\mathbf{W_{Skill}} \in \mathbb{R}^{T \times S}$. This matrix facilitates the soft partitioning of general skills. Within this framework, each element $w_{i}^{t}$ in $\mathbf{W_{Skill}}$ is intended to be a binary value, designating the activation of the adapter module $\phi_{i}$ for a specific weighted task set $t$. However, binary matrices such as $\mathbf{W_{Skill}}$ are non-differentiable, which precludes the application of gradient-based learning methods.

To circumvent this impediment, we employ the Gumbel sigmoid technique, as proposed by Maddison et al.\cite{maddison2016concrete} and Jang et al.\cite{jang2016categorical}, which facilitates the emulation of a set of continuously relaxed Bernoulli distributions. This method ensures both an element of stochasticity and the capability for diverse sampling:
\begin{equation}
\hat{w}_{i}^{t} = \sigma \left( \log \left[ \frac{\sigma(w)_{i}^{t}) u}{(1 - \sigma(w_{i}^{t}))(1 - u)} \right] \right), \quad u \sim \mathcal{U}(0,1)
\end{equation}
where $\sigma$ denotes the sigmoid function, and $u$ is drawn from a uniform distribution $\mathcal{U}(0,1)$.

\subsection{Parameter Efficiency}

We have achieved an efficient parameterization of skill modules by utilizing low-rank techniques. Each adapter in our $\texttt{OrchMoE}$ experiments is a Low-Rank Adapter (LoRA) as introduced by Hu et al.\cite{hu2022lora}. LoRA represents a simple yet potent architecture designed specifically for Transformer-based models, as described by Vaswani et al.\cite{10.5555/3295222.3295349}. The fundamental concept of LoRA is to decompose the weight matrix associated with the linear transformations in Transformers into the product of two low-rank matrices. Concretely, a linear projection $f : \mathbb{R}^{d} \rightarrow \mathbb{R}^{d}$ can be reformulated, excluding the bias term, as:

\begin{equation}
\mathbf{h}_{l+1} = \mathbf{h}_{l} \left[ \mathbf{W}_l + \Delta\mathbf{W} \right] = \mathbf{h}_{l} \left[ \mathbf{W}_l + \mathbf{W}_{\text{down}} \mathbf{W}_{\text{up}} \right]
\end{equation}

Rather than employing the full-rank update matrix $ \Delta\mathbf{W} \in \mathbb{R}^{d \times d}$, the model leverages two significantly smaller matrices, $\mathbf{W}_{\text{down}} \in \mathbb{R}^{d \times r}$ and $\mathbf{W}_{\text{up}} \in \mathbb{R}^{r \times d}$, where $r \ll d$, which are optimized via gradient descent.

By incorporating LoRA, the update for each linear layer in the model requires merely $2rd$ parameters to compute, in contrast to the $d^2$ parameters required originally. This shift markedly boosts parameter efficiency and fosters expedited training times, even on constrained computational resources.

As evidenced in Hu et al.~\cite{hu2022lora}, LoRA can be seamlessly integrated into various components of Transformer models, including the query, key, value, and feed-forward layers. Interestingly, the selection of the rank ( r ) is not critically constrained, underscoring LoRA's adaptability. For our experiments, we have replaced all query, key, and value layers with $\texttt{OrchMoE}$, constituting an amalgamation of multiple LoRA elements.

\section{Empirical Experiments}
\subsection{Experimental Setup}

To assess the efficacy of integrating multiple adapters into our PEFT framework $\texttt{OrchMoE}$, we conducted comprehensive experiments on the Super Natural Instructions (Super NI) dataset~\cite{wang-etal-2022-super}, which serves as a meta-dataset~\cite{DBLP:journals/corr/abs-1903-03096}. This dataset encompasses 76 distinct task types in the field of natural language processing, featuring over 1600 diverse NLP tasks. In our experimental setup, we randomly selected 10 and 100 tasks, and from each task, we randomly chose 1000 samples for training and a separate set of 100 samples for evaluation purposes. To evaluate the model's transferability to novel tasks, we further randomly selected an additional 10 tasks, for which 100 samples per task were randomly chosen for evaluation. To ensure comparability, our sampling approach adheres to the identical sampling methodology described in~\cite{ponti2022combining}. To gauge the performance of the trained model, we utilized a variety of metrics across all selected tasks, including Exact Match (EM) and Rouge metrics~\cite{lin2004rouge}, specifically Rouge-1, Rouge-L, and Rouge-LSum.

To confirm the universal applicability of our multitasking learning method $\texttt{OrchMoE}$, we selected T5 Version 1.1 - LM Adapted (T5)\cite{raffel2020exploring} and GLM\cite{du2021glm} as our base models. These models represent the Encoder-Decoder and Decoder-Only architectures of Transformer-based Large Language Models (LLMs), respectively.

In our research, we conducted thorough comparisons between our proposed approach, $\texttt{OrchMoE}$, and several existing methods akin to the Mixture of Experts (MoE) within the PEFT paradigm. The methodologies we benchmarked against include LoRA, MoE-LoRA, Poly, and MHR. This comparative analysis enables us to scrutinize and ascertain the relative performance and effectiveness of our framework in relation to these established PEFT methods.

\subsection{Training Details}
In our experimentation, the PEFT methodology was applied to the query, key, and value projections within all Transformer layers of the base model. This application was conducted to maintain a consistent number of trainable parameters across all evaluated methods. Distinct parameters were configured for each method to facilitate an equitable comparison. Specifically, for the vanilla LoRA implementation, we designated the rank $ r $ of the low-rank approximation to be 32. In the case of MoE LoRA, Poly, and MHR, a total of 4 parallel LoRA modules (experts) were employed, each with a reduced rank of $r = 8$. Regarding $\texttt{OrchMoE}$, we arranged a total of 4 parallel LoRA modules as skills, selecting a rank of $r = 4$ for each and the quantity of abstract tasks corresponds directly to the number of distinct tasks handled by the model.

During the training phase, we utilized a batch size of 4 on the Super Natural Instructions (Super NI) dataset and employed the AdamW optimizer~\cite{loshchilov2017decoupled} with an initial learning rate of $5 \times 10^{-5}$. Additionally, we adopted a linear decay learning rate strategy~\cite{loshchilov2016sgdr}, featuring a weight decay factor of 0.01 and a warmup ratio of 0.06. All experiments were carried out on a single NVIDIA Tesla A100 graphics processing unit.

\subsection{Main Results and Discussion}
\subsubsection{Comparative Analysis of Model Performance Across Task Scales}
\label{sec:task_scale_analysis}
We assessed the performance of diverse Parameter-Efficient Fine-Tuning (PEFT) methods by conducting two separate experiments on the Super Natural Instructions (SuperNI) dataset, all experiments are trained 1 epoch. In the first experiment, we evaluated the methods using a subset of 10 randomly selected tasks. Subsequently, in the second experiment, we expanded the scope to include 100 randomly selected tasks. These experiments were carried out across two different model architectures: T5-XXL, an Encoder-Decoder model with 11 billion parameters, and GLM-10B, a Decoder-Only model with 10 billion parameters. For each of these model architectures, we reported the average performance metrics of the PEFT methods on both the 10-task and 100-task experiments in table~\ref{tab:10B_model_compare_cross_tasks}.

The experimental results demonstrate that $\texttt{OrchMoE}$ is efficacious across models with varying architectures, notably surpassing other PEFT methods while maintaining an equivalent number of parameters. Unlike MoE-LoRA, $\texttt{OrchMoE}$ incorporates Task and Skill Routers, which augment the adaptability of parameters, thereby facilitating more effective task and skill acquisition. In contrast to methods such as Poly and MHR, which necessitate the explicit specification of task identifiers, $\texttt{OrchMoE}$ operates without requiring predefined task IDs. This attribute underscores its superior universality and contributes to its enhanced performance in the final evaluations.

The experimental outcomes across different task quantities reveal that $\texttt{OrchMoE}$ sustains its effectiveness even as the number of downstream tasks markedly escalates. This robustness is attributed to the ability of $\texttt{OrchMoE}$ to capture task-specific nuances through abstract tasks and skills while simultaneously sharing generalized abstract tasks and skills effectively. As the task count grows, the high parameter generalization capability of $\texttt{OrchMoE}$ becomes increasingly pronounced, leading to more substantial and consistent enhancements in performance.

\begin{table*}[htbp]
\centering
\resizebox{\textwidth}{!}{%
\begin{tabular}{@{}cllllllll@{}}
\toprule
\multicolumn{1}{l}{\multirow{2}{*}{Base Model}} &
  \multicolumn{4}{c}{Train and Testing on 10 Tasks} &
  \multicolumn{4}{c}{Train and Testing on 100 Tasks} \\ \cmidrule(l){2-9} 
\multicolumn{1}{l}{} &
  PEFT Type &
  Rouge1 &
  RougeL &
  \multicolumn{1}{l|}{RougeLsum} &
  PEFT Type &
  Rouge1 &
  RougeL &
  RougeLsum \\ \midrule
\multirow{5}{*}{T5-XXL} &
  LoRA &
  48.07 &
  47.87 &
  \multicolumn{1}{l|}{47.98} &
  LoRA &
  49.97 &
  48.89 &
  48.93 \\
 &
  MoE-LoRA &
  52.88 &
  52.51 &
  \multicolumn{1}{l|}{52.73} &
  MoE-LoRA &
  52.14 &
  51.12 &
  51.15 \\
 &
  Poly &
  54.62 &
  54.43 &
  \multicolumn{1}{l|}{54.51} &
  Poly &
  55.42 &
  54.65 &
  54.64 \\
 &
  MHR &
  55.00 &
  54.79 &
  \multicolumn{1}{l|}{54.88} &
  MHR &
  55.81 &
  55.01 &
  55.01 \\
 &
  \textbf{OrchMoE} &
  \textbf{61.43} &
  \textbf{61.31} &
  \multicolumn{1}{l|}{\textbf{61.24}} &
  \textbf{OrchMoE} &
  \textbf{69.96} &
  \textbf{69.09} &
  \textbf{69.14} \\ \midrule
\multirow{5}{*}{GLM-10B} &
  LoRA &
  41.58 &
  41.52 &
  \multicolumn{1}{l|}{41.67} &
  LoRA &
  50.02 &
  48.61 &
  48.58 \\
 &
  MoE-LoRA &
  40.99 &
  40.91 &
  \multicolumn{1}{l|}{40.93} &
  MoE-LoRA &
  50.54 &
  49.18 &
  49.17 \\
 &
  Poly &
  45.09 &
  44.97 &
  \multicolumn{1}{l|}{45.16} &
  Poly &
  56.02 &
  54.87 &
  54.88 \\
 &
  MHR &
  45.06 &
  45.08 &
  \multicolumn{1}{l|}{45.19} &
  MHR &
  56.47 &
  55.42 &
  55.44 \\
 &
  \textbf{OrchMoE} &
  \textbf{61.51} &
  \textbf{61.44} &
  \multicolumn{1}{l|}{\textbf{61.48}} &
  \textbf{OrchMoE} &
  \textbf{60.12} &
  \textbf{59.04} &
  \textbf{59.07} \\ \bottomrule
\end{tabular}%
}
\caption{T5-XXL and GLM-10B with different PEFT methods on the 10 and 100 randomly selected tasks from SuperNI dataset. We report the average Rouge1, RougeL, and RougeLsum for corresponding tasks. Higher is better for all metrics.}
\label{tab:10B_model_compare_cross_tasks}
\end{table*}

\subsubsection{Transfer Learning Analysis on Unseen Tasks}
\label{sec:transfer_learning_analysis}
The ability to transfer knowledge to unseen new tasks is a critical measure of multi-task learning methodologies. For our experiments, we randomly selected 100 tasks from the Super Natural Instructions (SuperNI) dataset for the training phase. A distinct set of 10 new tasks, also randomly chosen from the SuperNI dataset, was used for the evaluation phase to assess the generalization capabilities of the models. We conducted these evaluations on two different model architectures: T5 and GLM. All experiments were performed with the models being trained for a single epoch.

The experimental findings in table~\ref{tab:10B_model_compare_new_tasks} affirm that $\texttt{OrchMoE}$ substantially outperforms parameters comparable PEFT methods in terms of efficiently transferring existing knowledge to novel tasks, while also demonstrating strong advantages across various model architectures. When juxtaposed with the MoE LoRA results, it becomes evident that the nonlinear Router within $\texttt{OrchMoE}$ significantly bolsters the generalization efficacy of multiple adapters. Furthermore, in contrast to Poly and MHR, $\texttt{OrchMoE}$ leverages abstract Tasks and Skills to further amplify its parameter generalization capabilities beyond the limitations of predefined task categories. This approach enables the model to more effectively assimilate information from extant task samples, thereby translating it into valuable parameter knowledge for newly encountered downstream tasks.
\begin{table}[htbp]
\centering
\resizebox{0.5\textwidth}{!}{%
\begin{tabular}{@{}cllll@{}}
\toprule
\multicolumn{1}{l}{\multirow{2}{*}{Base Model}} & \multicolumn{4}{c}{Train on 100 Tasks and Testing on 10 new Tasks} \\ \cmidrule(l){2-5} 
\multicolumn{1}{l}{}                          & PEFT Type          & Rouge1         & RougeL         & RougeLsum      \\ \midrule
\multicolumn{1}{c|}{\multirow{5}{*}{T5-XXL}}  & LoRA               & 38.94          & 34.25          & 34.30          \\
\multicolumn{1}{c|}{}                         & MoE-LoRA           & 40.55          & 35.77          & 35.79          \\
\multicolumn{1}{c|}{}                         & Poly               & 43.19          & 38.29          & 38.30          \\
\multicolumn{1}{c|}{}                         & MHR                & 43.40          & 38.49          & 38.49          \\
\multicolumn{1}{c|}{}                         & \textbf{OrchMoE} & \textbf{54.52} & \textbf{48.40} & \textbf{48.47} \\ \midrule
\multicolumn{1}{c|}{\multirow{5}{*}{GLM-10B}} & LoRA               & 28.40          & 24.33          & 24.31          \\
\multicolumn{1}{c|}{}                         & MoE-LoRA           & 29.92          & 25.38          & 25.40          \\
\multicolumn{1}{c|}{}                         & Poly               & 29.07          & 23.79          & 23.80          \\
\multicolumn{1}{c|}{}                         & MHR                & 24.84          & 21.09          & 21.16          \\
\multicolumn{1}{c|}{}                         & \textbf{OrchMoE} & \textbf{31.62} & \textbf{27.59} & \textbf{27.61} \\ \bottomrule
\end{tabular}%
}
\caption{T5-XXL and GLM-10B with different PEFT methods on the 100 randomly selected tasks from SuperNI dataset. We report the average Rouge1, RougeL, and RougeLsum for 10 new tasks. Higher is better for all metrics.}
\label{tab:10B_model_compare_new_tasks}
\end{table}

\subsubsection{Comparative Performance Analysis of $\texttt{OrchMoE}$ Across Model Scales}
\label{sec:model_scale_compare}
To examine the effects of $\texttt{OrchMoE}$ and other PEFT methods on training efficiency across models of varying scales, we implemented a series of experiments using the Super Natural Instructions (SuperNI) dataset with two variants of the GLM model: GLM-10B and GLM-2B. In the first experiment, we assessed these methods on a subset of 10 tasks selected at random. The investigation was subsequently extended in a second experimental phase, where the analysis included an expanded set of 100 tasks, also chosen randomly. Given that the GLM-2B model demonstrated a requirement for additional epochs to achieve convergence, we standardized the training duration across all experiments to span 10 epochs.

The experimental results in table~\ref{tab:GLM_model_compare_scales} demonstrate a notable trend: models with a smaller number of parameters exhibit more significant improvements when employing $\texttt{OrchMoE}$. Remarkably, GLM-2B models that underwent $\texttt{OrchMoE}$ showed performance levels nearing those of the substantially larger GLM-10B counterparts. This phenomenon underscores the efficacy of $\texttt{OrchMoE}$'s design, which leverages abstract tasks and skills to substantially enhance parameter efficiency. The method of abstract information extraction and integration through information-crossing routers effectively raises the performance threshold for models with a limited number of parameters.

\begin{table*}[htbp]
\centering
\resizebox{\textwidth}{!}{%
\begin{tabular}{@{}c|llllllll@{}}
\toprule
\multicolumn{1}{l|}{\multirow{2}{*}{Base Model}} &
  \multicolumn{4}{c}{Train and Testing on 10 Tasks} &
  \multicolumn{4}{c}{Train and Testing on 100 Tasks} \\ \cmidrule(l){2-9} 
\multicolumn{1}{l|}{} &
  PEFT Type &
  Rouge1 &
  RougeL &
  \multicolumn{1}{l|}{RougeLsum} &
  PEFT Type &
  Rouge1 &
  RougeL &
  RougeLsum \\ \midrule
\multirow{5}{*}{GLM-2B} &
  LoRA &
  27.60 &
  27.45 &
  \multicolumn{1}{l|}{27.49} &
  LoRA &
  52.29 &
  51.53 &
  51.56 \\
 &
  MoE-LoRA &
  55.65 &
  55.30 &
  \multicolumn{1}{l|}{55.37} &
  MoE-LoRA &
  53.35 &
  52.51 &
  52.57 \\
 &
  Poly &
  54.17 &
  55.23 &
  \multicolumn{1}{l|}{55.28} &
  Poly &
  57.11 &
  56.40 &
  56.31 \\
 &
  MHR &
  55.07 &
  56.08 &
  \multicolumn{1}{l|}{56.25} &
  MHR &
  58.06 &
  57.26 &
  57.30 \\
 &
  \textbf{OrchMoE} &
  \textbf{64.16} &
  \textbf{65.27} &
  \multicolumn{1}{l|}{\textbf{65.46}} &
  \textbf{OrchMoE} &
  \textbf{64.18} &
  \textbf{63.33} &
  \textbf{63.33} \\ \midrule
\multirow{5}{*}{GLM-10B} &
  LoRA &
  61.69 &
  61.66 &
  \multicolumn{1}{l|}{61.77} &
  LoRA &
  61.34 &
  60.24 &
  60.26 \\
 &
  MoE-LoRA &
  62.47 &
  62.52 &
  \multicolumn{1}{l|}{62.62} &
  MoE-LoRA &
  61.04 &
  59.91 &
  59.95 \\
 &
  Poly &
  61.87 &
  61.99 &
  \multicolumn{1}{l|}{61.94} &
  Poly &
  64.55 &
  63.41 &
  63.26 \\
 &
  MHR &
  62.90 &
  62.94 &
  \multicolumn{1}{l|}{63.03} &
  MHR &
  65.62 &
  64.38 &
  64.37 \\
 &
  \textbf{OrchMoE} &
  \textbf{66.99} &
  \textbf{67.04} &
  \multicolumn{1}{l|}{\textbf{67.15}} &
  \textbf{OrchMoE} &
  \textbf{67.48} &
  \textbf{66.29} &
  \textbf{66.27} \\ \bottomrule
\end{tabular}%
}
\caption{GLM-2B and GLM-10B with different PEFT methods on the 10 and 100 randomly selected tasks from SuperNI dataset. We report the average Rouge1, RougeL, and RougeLsum for corresponding tasks. Higher is better for all metrics.}
\label{tab:GLM_model_compare_scales}
\end{table*}

\subsubsection{Parameter Efficiency Analysis}
\label{sec:parameter_efficiency}
\begin{figure}[htbp]
    \centering
    \includegraphics[width=0.5\textwidth]{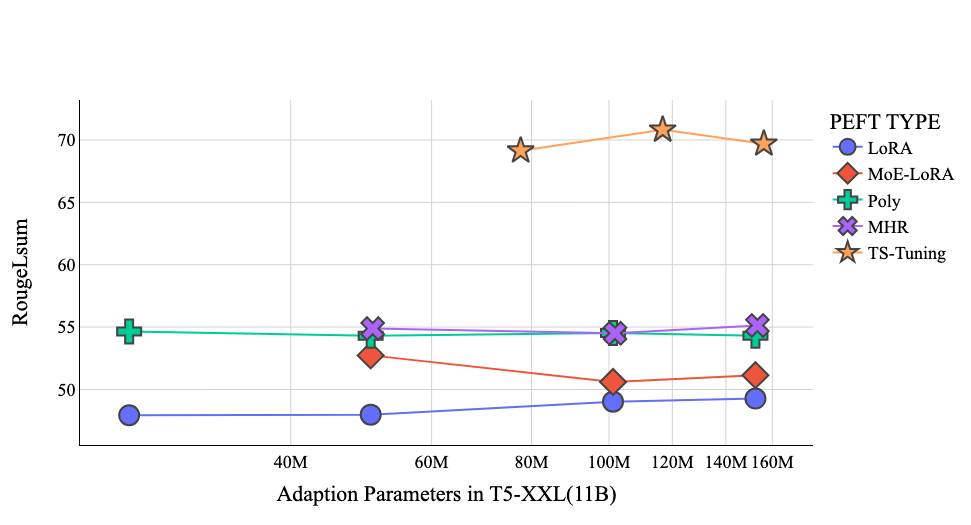}
    \caption{RougeLsum of PEFT methods on SuperNI 100 Tasks dataset when applied on T5-XXL. The X-axis shows the trainable parameter count during the fine-tuning process.}
    \label{fig:autotask_parameter_efficiency}
\end{figure}

Figure~\ref{fig:autotask_parameter_efficiency} presents an evaluation of the effect of training parameter quantity on the final performance for each PEFT method when trained for a single epoch using the T5-XXL model. In this assessment, the adapter consistently employs LoRA, and by varying the rank size of LoRA, we can assess the influence of different parameter quantities within the same PEFT method on overall performance, as measured by the RougeLsum metric.

Among the PEFT methods evaluated at the same parameter scale, $\texttt{OrchMoE}$ stands out with exceptional performance. Notably, $\texttt{OrchMoE}$ configurations with fewer parameters consistently surpass competing PEFT methods that utilize larger parameter sizes. Moreover, our analysis suggests that the impact of parameter size on the performance of various PEFT methods is marginal, aligning with the conclusions drawn by the original developers of LoRA. These results underscore the efficacy of the Router, which is structured around the principles of abstract Tasks and Skills, enabling $\texttt{OrchMoE}$ to achieve superior parameter learning efficiency. As such, $\texttt{OrchMoE}$ emerges as a powerful paradigm for optimizing parameter efficiency in multi-task adaptation scenarios.

\subsubsection{In-Depth Analysis of Learned Task and Skills}
As outlined in the former sections, $\texttt{OrchMoE}$ showcases significant advantages across diverse architectures, task scales, model sizes, and parameter efficiency metrics. In this section, we delve into the intrinsic characteristics of $\texttt{OrchMoE}$ that underpin these performance enhancements.

\begin{figure}[h]
    \centering
    \includegraphics[width=0.5\textwidth]{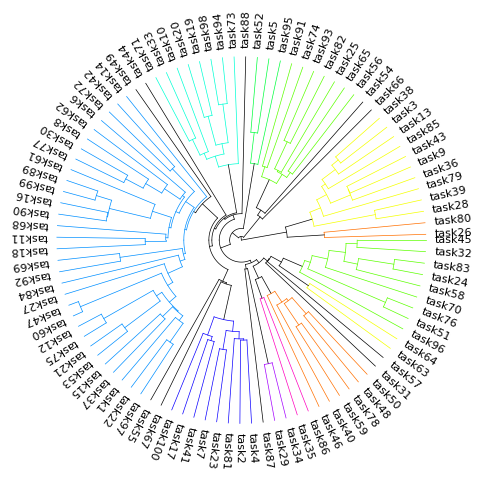}
    \caption{Task clustering dendrogram for Task-skill allocation matrix $W$ of $\texttt{OrchMoE}$ using GLM-10B as the base model, set 100 abstract tasks in NI-100-Tasks experiment.
Tasks are grouped into the same category if they share a similar subset of skills.}
    \label{fig:GLM_10B_TS_analysis_100Task}
\end{figure}

To gain a deeper understanding of the learning dynamics and parameter efficiency inherent in $\texttt{OrchMoE}$, we refer to Figure~\ref{fig:GLM_10B_TS_analysis_100Task}, which illustrates the allocation matrix for the abstract 'Task' and abstract 'skill' within the Skill Router of $\texttt{OrchMoE}$, utilizing the GLM-10B as the foundational model in the SuperNI-100 experiment. Following proper normalization, these allocation matrices uncover notable disparities in skill distribution across varying tasks. Moreover, we embarked on task clustering by analyzing the skill assignments acquired through all layers of the GLM-10B model, which has been trained on the SuperNI-100 dataset. The emergent patterns indicate that the abstract tasks displayed a heightened capacity for abstraction and effectively segmented the collection of 100 abstract tasks into distinct groupings. 

The abstract task clustering results reveal that many tasks share commonalities in skill allocation, thereby indicating a potential for information redundancy within the system. $\texttt{OrchMoE}$ is designed to efficiently distill and transmit information across actual tasks through its abstract tasks. An experiment to reduce the number of these abstract tasks yielded insightful findings, as depicted in table~\ref{tab:various_abstract_tasks_compare}. Despite a slight decrease in performance, the outcomes demonstrate the model's robustness and stability, suggesting that $\texttt{OrchMoE}$ possesses a significant degree of knowledge tolerance. This attribute can be ascribed to the model's pronounced ability to abstractly represent tasks and its potent capacity for knowledge transfer across tasks with similar characteristics.

\begin{table*}[htbp]
% \small
\centering
\resizebox{\textwidth}{!}{%
\begin{tabular}{@{}llllllll@{}}
\toprule
\multirow{2}{*}{\begin{tabular}[c]{@{}l@{}}OrchMoE \\ Abstract Task Setting\end{tabular}} &
  \multicolumn{1}{l|}{\multirow{2}{*}{\begin{tabular}[c]{@{}l@{}}Adapter-to-Base \\ Parameter Ratio\end{tabular}}} &
  \multicolumn{3}{c|}{train and testing on 100 tasks} &
  \multicolumn{3}{c}{\begin{tabular}[c]{@{}c@{}}train on 100 tasks and \\ testing on 10 new tasks\end{tabular}} \\ \cmidrule(l){3-8} 
       & \multicolumn{1}{l|}{} & rouge1 & rougeL & \multicolumn{1}{l|}{rougeLsum} & rouge1 & rougeL & rougeLsum \\ \midrule
100 & 0.39\%                & 60.12  & 59.04  & 59.07                          & 31.62  & 27.59  & 27.61     \\
80  & 0.35\%                & 60.25  & 59.19  & 59.20                          & 31.50  & 27.86  & 27.84     \\
60  & 0.31\%                & 59.88  & 58.86  & 58.87                          & 31.63  & 27.57  & 27.58     \\
40  & 0.27\%                & 60.08  & 59.06  & 59.06                          & 32.13  & 27.96  & 27.98     \\
20  & 0.23\%                & 59.54  & 58.45  & 58.49                          & 31.95  & 27.85  & 27.86     \\ \bottomrule
\end{tabular}%
}
\caption{$\texttt{OrchMoE}$ Experiments: GLM-10B trained 1 epoch on SuperNI-100 with Varied Abstract Task Settings}
\label{tab:various_abstract_tasks_compare}
\end{table*}

\section{Related Works}
\subsection{Large Language Model}
% Research in language modeling, crucial for both understanding and generation tasks, has undergone significant transformation over the past two decades. The field has evolved from statistical language models to advanced neural networks, culminating in the recent advent of pre-trained language models (PLMs). These PLMs are based on the Transformer architecture and are pre-trained on extensive text corpora, displaying remarkable effectiveness across a spectrum of natural language processing tasks. A notable trend in the field is model upscaling; researchers have observed that increasing the number of parameters can yield better performance. Intriguingly, once these models surpass a certain threshold in parameter count, they not only continue to exhibit performance gains but also begin to exhibit in-context learning capabilities, which are absent in their smaller counterparts. This has led the research community to coin the term 'Large Language Model' (LLM) to describe PLMs of substantial size—those with hundreds of millions, or even billions, of parameters.

The Transformer~\cite{vaswani2023attention} architecture constitutes the bedrock upon which the preponderance of contemporary large-scale models are constructed. Stemming from this versatile architecture, two principal model frameworks have emerged: the Encoder-Decoder and the Decoder-Only architectures. The Encoder-Decoder paradigm, exemplified by Google's T5 model~\cite{raffel2023exploring}, has been celebrated for its proficiency in semantic comprehension, attributable to its cloze-style learning objectives. However, this architecture also exhibits certain limitations in generative tasks, an issue that becomes more pronounced in Encoder-Only models such as BERT~\cite{devlin2019bert}. Conversely, Decoder-Only models, epitomized by GPT~\cite{radford2019language} and GLM~\cite{du2022glm}, have demonstrated formidable strengths in both natural language generation and in-context learning. Their prowess is largely ascribed to the uniformity of autoregressive generation during both training and inference phases, offering a more coherent framework for these tasks.

\subsection{Parameter Efficient Tuning}
To mitigate the high costs of training large Transformer models, Parameter Efficient Fine-tuning (PEFT) employs adapters—small, trainable modules inserted between existing layers. PEFT updates only these adapter weights, keeping the main model's parameters fixed, thus reducing computational burden. A succession of fundamental adapter methodologies has been proposed, among which Layerwise Optimized Rank Adaptation (LoRA)~\cite{DBLP:journals/corr/abs-2106-09685} exerts a prominent influence. LoRA endeavors to augment parameter efficiency through the application of low-rank decomposition, fine-tuning the model by optimizing low-rank matrix weights in synchronization with their associated sovereignty matrices. Innovations such as (IA)$^3$~\cite{liu2022few} build upon the principles of LoRA by moderating and intensifying the internal activations through injection adapters and employing learned vectors for the recalibration of these internal activations. Furthermore, an array of methods including P-tuning~\cite{liu2022ptuning}, Prompt tuning~\cite{lester2021power}, and Prefix tuning~\cite{li2021prefixtuning} have been successively introduced, signifying an emergent trajectory in the tuning paradigms tailored to the domain of large-scale models.

\section{Conclusion}

In this work, we introduce a multi-faceted fine-tuning paradigm predicated on Parameter Efficient Fine-tuning (PEFT), which reconceptualizes the model's multi-task learning into dual aspects—Task learning and Skill learning, collectively named $\texttt{OrchMoE}$. Through the innovative deployment of 'Task Router' and 'Skill Router' structures, our approach facilitates efficient multi-task fine-tuning on expansive language models, even under the constraint of limited computational resources. We engage in an in-depth exploration of the TS Training mechanism and present a comprehensive comparative analysis of its performance benchmarks. TS Training exhibits exceptional proficiency, outstripping extant PEFT baselines and setting new records for state-of-the-art performance by leveraging its remarkable parameter efficiency coupled with superior multi-task abstraction and migration capabilities. These findings underscore the transformative potential and import of our multi-task PEFT fine-tuning approach.
% Acknowledgements are optional. In the camera-ready version you may include an unnumbered acknowledgments section, including acknowledgments of help from colleagues, financial support, and permission to publish. This is not allowed in the anonymous submission. If present, acknowledgements must be in a dedicated, unnumbered section appearing after all regular sections but before references.  This section may be placed on the References pages.

% Use
% \begin{quote}
%     {\tt \textbackslash{}section*\{Acknowledgements\}}
% \end{quote}
% to typeset the acknowledgements section in \LaTeX{}.

\appendix

% \section*{Ethical Statement}

% There are no ethical issues.

% \section*{Acknowledgments}

% The preparation of these instructions and the \LaTeX{} and Bib\TeX{}
% files that implement them was supported by Schlumberger Palo Alto
% Research, AT\&T Bell Laboratories, and Morgan Kaufmann Publishers.
% Preparation of the Microsoft Word file was supported by IJCAI.  An
% early version of this document was created by Shirley Jowell and Peter
% F. Patel-Schneider.  It was subsequently modified by Jennifer
% Ballentine, Thomas Dean, Bernhard Nebel, Daniel Pagenstecher,
% Kurt Steinkraus, Toby Walsh, Carles Sierra, Marc Pujol-Gonzalez,
% Francisco Cruz-Mencia and Edith Elkind.

\bibliographystyle{named}
\bibliography{ijcai23}

\end{document}